\def\year{2021}\relax
\documentclass[letterpaper]{article} 
\usepackage{aaai21}  
\usepackage{times}  
\usepackage{helvet} 
\usepackage{courier}  
\usepackage[hyphens]{url}  
\usepackage{graphicx} 
\usepackage[switch]{lineno}  %
\usepackage{natbib}  
\usepackage{caption} 
\title{Dual-Branch Network with Dual-Sampling Modulated Dice Loss \\for Hard Exudate Segmentation from Colour Fundus Images}
\author{
    Qing Liu\textsuperscript{\rm 1}, Haotian Liu\textsuperscript{\rm 1}, Yixiong Liang \textsuperscript{\rm 1}
    \\
}
\affiliations{

    \textsuperscript{\rm 1} School of Computer Science, Central South University\\

}

\begin{document}
\maketitle

\begin{abstract}
Automated segmentation of hard exudates in colour fundus images is a challenge task due to issues of extreme class imbalance and enormous size variation. This paper aims to tackle these issues and proposes a dual-branch network with dual-sampling modulated Dice loss. It consists of two branches: large hard exudate biased segmentation branch and small hard exudate biased segmentation branch. Both of them are responsible for their own duties separately. Furthermore, we propose a dual-sampling modulated Dice loss for the training such that our proposed dual-branch network is able to segment hard exudates in different sizes. In detail, for the first branch, we use a uniform sampler to sample pixels from predicted segmentation mask for Dice loss calculation, which leads to this branch naturally be biased in favour of large hard exudates as Dice loss generates larger cost on misidentification of large hard exudates than small hard exudates. For the second branch, we use a re-balanced sampler to oversample hard exudate pixels and undersample background pixels for loss calculation. In this way, cost on misidentification of small hard exudates is enlarged, which enforces the parameters in the second branch fit small hard exudates well. Considering that large hard exudates are much easier to be correctly identified than small hard exudates, we propose an easy-to-difficult learning strategy by adaptively modulating the losses of two branches. We evaluate our proposed method on two public datasets and results demonstrate that ours achieves state-of-the-art performance.
\end{abstract}

\section{Introduction}
\label{sec1}
\noindent Hard exudate is one of the most significant manifestation of diabetic retinopathy (DR) \cite{klein1987wisconsin}. Automated and accurate segmentation of hard exudate in colour fundus images has several potential applications in clinical such as large-scale automated DR screening, computer-aided diagnosis and severity level assessment of DR \cite{sasaki2013quantitative}. 


The segmentation of hard exudate can be formulated as a dense classification problem.  At the era of deep learning, without any exception, the first choice is fully convolutional networks (FCNs). Its goal is to optimise parameters in designed FCNs to best fit the exudate ground-truth via minimising a specified loss function. However, achieving this goal is challenge due to two issues: 
\begin{itemize}
	\item \textbf{ Extreme class imbalance.} To illustrate how extreme the class imbalance is, we account the ratio of negative samples (i.e. background pixels) to positive samples (i.e. hard exudate pixels) in two public datasets for hard exudate segmentation, i.e.  DDR \cite{DDR} and IDRiD \cite{idrid2018} (see Table. \ref{ratio}). The ratio in DDR \cite{DDR} and IDRiD \cite{idrid2018} is $512$ and $110$ respectively. With those serious extreme class imbalanced data, how to design loss function and train the segmentation model to alleviate the bias towards majority class becomes critical. 
	\item \textbf{Enormous variation in size across connected components of hard exudate regions.} Most of hard exudate connected regions are small. In particular, we calculate the relative area of connected hard exudate regions and find that almost $90\%$ hard exudate pixels belong to connected regions whose relative area to the whole fundus image is less than $9.7\times 10^{-5}$ in DDR \cite{DDR} and $1.1\times10^{-4}$ in IDRiD \cite{idrid2018}. More seriously, there are $10\%$ hard exudate pixels belonging to connected hard exudate regions with relative areas less than $2.0\times10^{-6}$ in DDR \cite{DDR} and $5.0\times 10^{-6}$ in IDRiD \cite{idrid2018} respectively. The size variations of largest $10\%$ and smallest $10\%$ in DDR \cite{DDR} and IDRiD \cite{idrid2018} are almost $48$ and $22$ times, respectively. This variation in size which the FCN model needs to handle is enormous and rises a huge challenge for representation and classifier learning.
\end{itemize}

\begin{table}[t]
	\caption{Class distribution imbalance and exudate region size variation existing in exudate segmentation datasets DDR \cite{DDR} and IDRiD \cite{idrid2018}. $Ratio_{neg/pos}$ denotes the ratio of background pixels and hard exudate pixels. $Size_{large}$ and $Size_{small}$ are the relative size to images of the top $10\%$ largest exudate regions and top $10\%$ smallest exudate regions in the whole dataset.}
	\centering
	\begin{tabular}{c|c|c}
		\hline Dataset &  $Ratio_{neg/pos}$ & $Size_{large}$/$Size_{small}$ \\ \hline
		DDR & 512 &  $9.7\times10^{-5}$/$2.0\times 10^{-6}$\\
		IDRiD & 110 & $1.1\times 10^{-4}$/$5.0\times 10^{-6}$ \\ \hline
	\end{tabular}
	
	\label{ratio}
\end{table}

\begin{figure*}[t]
	\centering
	\includegraphics[width=\textwidth]{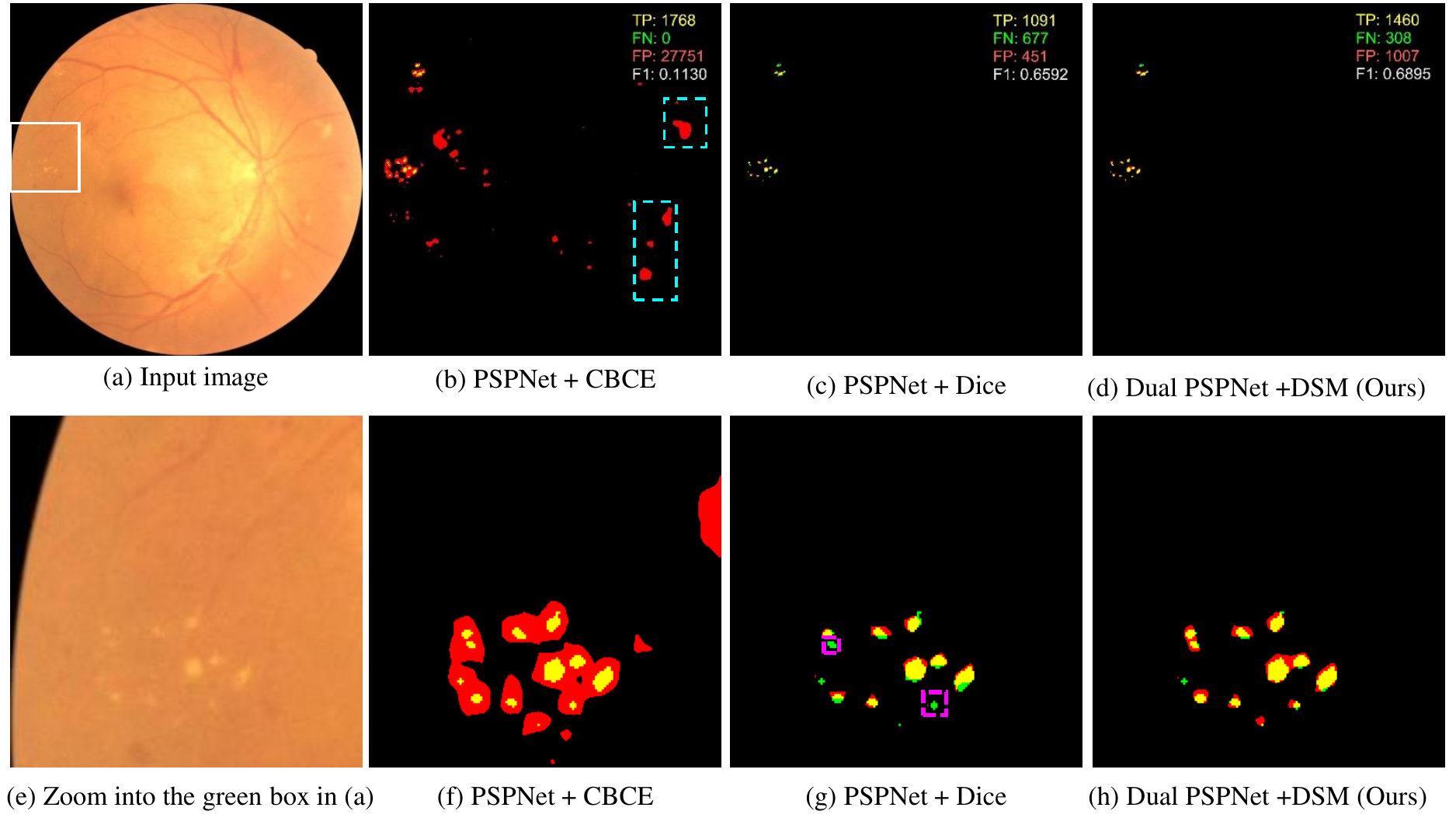} 
	\caption{Segmentation results by PSPNet \cite{PSPNet} trained with CBCE loss and Dice loss and our proposed dual-branch network. For better visualisation, we show the segmentation results of the entire image in (a) in the first row and zooming in the white solid window in (a) in the second row. Pixels in yellow are exudate pixels that are correctly classified. Pixels in red are background pixels that are wrongly classified as exudate pixels. Pixels in green are exudate pixels that are wrongly classified as background pixels. Dashed cyan box highlights background regions wrongly identified as exudate and dashed magenta box highlights exudate regions wrongly identified as background. From this figure we can see that PSPNet \cite{PSPNet} trained with CBCE loss tends to classify background as exudate while it trained with Dice loss tends to result in misidentification on hard exudate in small size. On the contrary, our dual-branch segmentation with DSM loss based on PSPNet \cite{PSPNet} achieves better than the single branch ones.}
	\label{eg4intro}
\end{figure*}

An intuitive way for hard exudate segmentation is to fine-tune semantic segmentation networks, such as HED \cite{HED}, PSPNet \cite{PSPNet}, Deeplabv3 \cite{Deeplabv3} and Deeplabv3+ \cite{Deeplabv3plus}, which are originally designed for dense classification tasks on natural scene images. Those methods handle the issue of class imbalance via using class balanced cross-entropy (CBCE) loss rather than the traditionally used cross-entropy loss. Inspired by HED \cite{HED}, Guo et al. \cite{guo2019l-seg, binloss} propose a variant of CBCE loss called bin loss and fine-tune the parameters in HED \cite{HED} for hard exudate segmentation. Bin loss \cite{guo2019l-seg, binloss} considers not only the class imbalance problem but also the hardness of background pixels to be correctly classified. FCNs trained with CBCE loss successfully avoid background bias. However, directly up-weighting loss of minority class and down-weighting loss of majority class according to the inverse class frequency is too rough. This makes the segmentation model be biased in favour of exudate pixels. Taking PSPNet \cite{PSPNet} as example, when training it with CBCE loss, the model always wrongly identifies confusion structures and background regions around hard exudates, as shown in Fig.~\ref{eg4intro} (b) and (f). 

An alternate strategy is to train FCNs with Dice loss \cite{diceloss}. Dice loss \cite{diceloss} is a regional loss which measures the overlapping error between the prediction and ground truth. It works better than CBCE loss when class imbalance issue is serious. However, because the costs on small exudate regions in terms of Dice loss are slight comparing to that on large hard exudate regions, FCNs trained with Dice loss is biased towards large hard exudate regions and results in dentification on small large hard exudates. The large variation in size across connected components of hard exudate makes the bias more serious. Fig. \ref{eg4intro} (c) and (g) show the results by PSPNet \cite{PSPNet} trained with Dice loss, which tends to mis-classifiy small hard exudate regions as background.

\begin{figure*}[t]
	\centering
	\includegraphics[width=\textwidth]{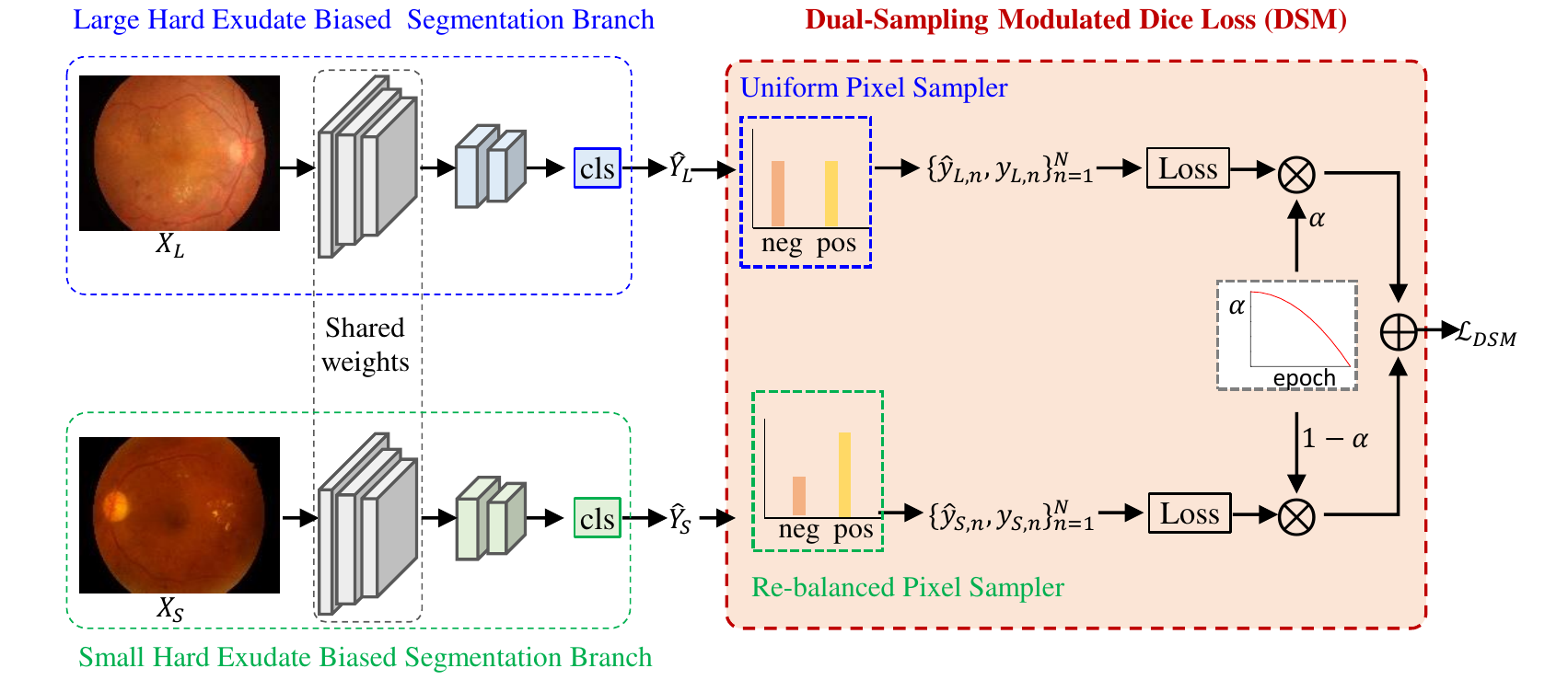} 
	\caption{Illustration of dual-branch network. The left part is the dual-branch network which is constructed with two identical segmentation branches with partial weight sharing. We note here arbitrary segmentation models such as PSPNet \cite{PSPNet}, Deeplabv3 \cite{Deeplabv3} and HED \cite{HED} are desired. The right part illustrates the proposed dual-sampling modulated Dice loss, which uses a uniform pixel sampler and a re-balanced pixel sampler sample pixels involving loss calculation. The two losses are adaptively modulated by a hyper-parameter which is set according to training epoch.}
	\label{framework}
\end{figure*}

In this paper, we propose a dual-branch network with dual-sampling modulated Dice loss to take care of both large and small hard exudate connected regions. As shown in Fig. \ref{framework}, our dual-branch network consists of two branches: large hard exudate biased segmentation branch and small hard exudate biased segmentation branch. It is trained with a dual-sampling modulated (DSM) Dice loss. Each branch separately performs its own duties for representation and classifier learning for hard exudates in different sizes. Large hard exudate biased segmentation branch learns a segmentation model which is large hard exudate biased while small hard exudate biased segmentation branch is biased in supporting small hard exudates. The bias of both learning branches is achieved by the proposed DSM loss. For large hard exudate biased segmentation branch, Dice loss with uniform pixel sampler is used. For small hard exudate biased segmentation branch, a re-balanced pixel sampler is used to oversample hard exudate pixels and undersample background pixels. Thus hard exudate pixels are sampled multiple times, which increases the penalty on misidentification of small hard exudate regions. In this way it well compensates the large hard exudate biased segmentation branch. The bias of two branches are shifted by modulation with regard to learning epoch. We evaluate the effectiveness of the proposed dual-branch network on DDR \cite{DDR} and IDRiD \cite{idrid2018} and results show that our dual-branch network outperforms existing hard exudate segmentation methods. Furthermore, to demonstrate the effectiveness of underlying thoughts of dual-branch network, we combine it with several dense classification networks. Results show that dual-branch networks trained with our proposed dual-sampling modulated Dice loss achieve superior performance to single branch networks trained with Dice loss. 

In summary, the contributions of this paper are as follows:
\begin{itemize}
	\item 	We propose a novel framework named dual-branch network to handle the issues of extreme class imbalance and enormous variation in size existed in the task of automated hard exudate segmentation from colour fundus images.
	\item  We propose a dual-sampling modulated Dice loss to guide the learning process of dual-branch network, which is an easy-to-difficult learning strategy and adaptively modulates the losses of two branches such that dual-branch network gradually shifts the attention to easy task of large hard exudate segmentation to hard task of small hard exudate segmentation.
	\item We conduct extensive experiments on two public datasets DDR \cite{DDR} and IDRiD \cite{idrid2018} and demonstrate that dual-branch network achieves state-of-the-art performance on hard exudate segmentation.
\end{itemize}

\section{Related Work}
\textbf{Unsupervised Hard Exudate Segmentation Methods.} Earlier methods such as \cite{walter2002contribution, sopharak2008automatic, ravishankar2009automated, welfer2010coarse} adopt morphological operations to enhance exudates, then use a simple thresholding to partition exudate from background. Similarly, Pereira et al. \cite{pereira2015exudate} propose to use median filtering and normalisation on green plane fundus image for enhancement. J. Kaur and D. Mittal \cite{kaur2018a} first remove the vessel and optic disc, then enhance the image by adaptive image quantization and normalisation. Finally, adaptive thresholding is used to identify exudates. Those kinds of methods are simple and do not need expensive annotations by ophthalmologists, but they always fail on confused structures which have high contrast to background.

\textbf{Coarse-to-fine Supervised Hard Exudate Segmentation Methods.} Those methods are data-driven and require expert annotation. They commonly involve two stages: (1) coarse detection stage for candidate detection and (2) fine segmentation stage to finally determine whether the candidate is hard exudate region. For example, in \cite{zhang2014exudate, wang2020hard}, candidates are extracted by mathematical morphology first, then a random forest is trained for classification. Rather than learning to determine whether the candidate is hard exudate, other researchers focus on high quality candidate extraction via learning. For example, Liu et al. \cite{liu2017location} first learn to extract multiscale hard exudate candidate patches and reduces numerous background regions, then identify hard exudate regions from candidate patches according to their characteristics such as intensity contrast to background and area. Kusakunniran et al. \cite{kusakunniran2018hard} first learn a multilayer perceptron for the detection of candidate hard exudate seeds. With the clusters of initial seeds, iterative graph-cut is used for segmentation. Additionally,  Parham et al. \cite{khojasteh2019novel, khojasteh2019exudate} propose to identify exudate patches from non-exudate patches by either training a lightweight deep network on candidate patches or training a support vector machine with features extracted by pre-trained deep network Resnet-50 \cite{ResNet}. However, how to extract hard exudate candidate patches from whole images during testing phase still needs to be solved.  

\textbf{End-to-end Hard exudate Segmentation Methods.} Recent hard exudate segmentation methods adopt an end-to-end manner to train an FCN with a loss function. Mo et al. \cite{mo2018exudate} design a fully convolutional residual network named FCRN for exudate segmentation while Guo et al. \cite{LWENet} design a lightweight neural network named LWENet. In \cite{guo2019l-seg}, L-seg is proposed for multi-lesion segmentation. All of those methods take the class imbalance problem into consideration and train the networks with CBCE loss. In \cite{binloss}, an improved CBCE loss incorporating hard negative mining is proposed for hard exudate segmentation. Both CBCE and bin loss avoid the background bias by increasing the cost of wrong classification on exudate pixels. However, due to the imbalanced cost weights, FCNs trained with CBCE loss and bin loss turn to suffer from exudate-bias.

It is noteworthy to mention that dual-network and dual-sampling have been used for class imbalance classification. In \cite{zhou2020bbn}, bilateral branch network (BBN) equipped with two samplers is proposed for class imbalance image classification. In \cite{ouyang2020dual}, a dual-sampling network (DSN) which consists of two separate branches with two samplers is proposed for diagnosis of COVID-19. Our method is inspired by BBN \cite{zhou2020bbn} and DSN \cite{ouyang2020dual}. Although all of those methods contain dual-branches with dual-samplers, ours differs from BBN \cite{zhou2020bbn} and DSN \cite{ouyang2020dual} in three aspects: (1) our dual-branch network is designed for dense classification while BBN \cite{zhou2020bbn} and DSN \cite{ouyang2020dual} are for image-level classification; (2) Images fed into 
dual-branch network are randomly sampled from training set while images fed into BBN \cite{zhou2020bbn} and DSN \cite{ouyang2020dual} are sampled according to the pre-defined samplers. (3) Samplers in dual-branch network are used on predicted segmentation masks which sample pixels involving Dice loss calculation while samplers in BBN \cite{zhou2020bbn} and DSN \cite{ouyang2020dual} are used in input layer which sample images for representation and classifier learning.


\section{Method}
\subsection{Overall Framework}
Our goal is to pursuit deep network parameters that best fit the hard exudate ground-truth in different sizes with given training images. For hard exudates in different sizes, learning representation and classifier in different manners is desired. To this end, we propose to adopt two branches to separately learn representation and classifier. One branch, named large hard exudate biased segmentation branch, is mainly responsible for hard exudates in large size. The other, named small hard exudate  biased segmentation branch, is responsible for hard exudates in small size. To achieve the size bias of each branch adaptively, we design a dual-sampling modulated Dice loss, termed DSM. Fig. \ref{framework} illustrates our proposed dual-branch network.

Formally, let $X\in \mathcal{R}^{H\times W\times3}$ denote a training colour fundus image size of $H\times W$ and $Y\in \mathcal{R}^{H\times W}$ is the corresponding ground-truth, which is a binary map within the context of hard exudate segmentation. From training set, we randomly fetch two images $\{X_L, Y_L\}$ and $\{X_S, Y_S\}$ and feed them into large hard exudate biased segmentation branch and small hard exudate biased segmentation branch respectively to obtain the final predictions $\hat{Y}_L$ and $\hat{Y}_S$. Next we elaborate the architecture of our dual-branch network and training details with our dual-sampling modulated Dice loss.

\subsection{Dual-Branch Segmentation Network}

We let both branches economically share the same segmentation network structure, as illustrated in Fig. \ref{framework}. Our dual-branch segmentation network can adopt arbitrary segmentation network. In this paper, we take three state-of-the-art segmentation networks, i.e. PSPNet \cite{PSPNet}, Deeplabv3 \cite{Deeplabv3} and HED \cite{HED} as examples to introduce our dual-branch segmentation network. PSPNet \cite{PSPNet} and Deeplabv3 \cite{Deeplabv3} adopt ResNet50 \cite{ResNet} equipped with dilation convolution \cite{Deeplab} in the last two stages as the backbone while HED \cite{HED} adopts the five-stage VGG16 \cite{VGGNet} as backbone. Both ResNet50 \cite{ResNet} and VGG16 \cite{VGGNet} contain five stages of convolutions. Additionally, in PSPNet \cite{PSPNet}, a pyramid pooling module is attached in the last convolutional stage, then a classifier is followed to make dense predictions. An auxiliary classifier is attached on the second convolution stage and auxiliary loss is generated to help optimise the learning process. In Deeplabv3 \cite{Deeplabv3}, an atrous spatial pyramid pooling module is attached on the last convolutional stage to generate multiscale feature maps. In HED \cite{HED}, five side-output layers are stitched on five convolutional stages and finally a fusion layer is used to aggregate the side-output predictions. To reduce the model complexity and speed up the inference, for PSPNet \cite{PSPNet} and Deeplabv3 \cite{Deeplabv3} as segmentation branch, weights in first four stages of backbone networks are shared while rest weights are learned separately. For HED \cite{HED}, only the first three stages of backbone networks are shared. In this way, the representations for final classifiers are specific to hard exudate in different sizes. The loss items in PSPNet \cite{PSPNet}, Deeplabv3 \cite{Deeplabv3} and HED \cite{HED} are replaced with our proposed dual-sampling modulated Dice loss.


\subsection{Loss Function}
In each training iteration, two pairs of fundus images and their ground truth denoted by $\{X_L, Y_L\}$ and $\{X_S, Y_S\}$ are randomly fetched. $X_L$ is fed into the large hard exudate biased segmentation branch and predictions are obtained and denoted by $\hat{Y}_L$. Similarly, $X_S$ is fed into the small hard exudate biased segmentation branch and predictions are obtained and denoted by $\hat{Y}_S$. As hard exudate segmentation suffers from the issues of extreme class imbalance and large variation in size, rather than using the class balanced cross-entropy loss \cite{guo2019l-seg, binloss}, we propose the dual-sampling modulated Dice loss (DSM loss). As shown in Fig. \ref{framework}, the total loss can be expressed as:
\begin{equation}
\mathcal{L}_{total} = \mathcal{L}_{DSM}\;.
\label{eq:largefinal}
\end{equation}

\textbf{Dual-Sampling Modulated Dice Loss.}  In our design of DSM loss, two different samplers are used to sample pixels from predictions by two branches separately. Then Dice loss is used to measure the dissimilarity between set of sampled pixels and their ground truths.

For large hard exudate biased segmentation branch, with predicted segmentation mask $\hat{Y}_L$, we use a uniform pixel sampler which samples hard exudate pixels and background pixels with equal probability. We denote the sampled pixel set with $\hat{\mathcal{S}}_{L}=\{\hat{y}_{L,n}\}_{n=1}^{N}$ where $N=H\times W$ and the corresponding set of ground truth with $\mathcal{S}_{L}=\{y_{L,n}\}_{n=1}^{N}$ where $y_{L, n}$ is ground truth of $\hat{y}_{L, n}$. We use Dice loss \cite{diceloss} to calculate the dissimilarity between sampled pixel set and the corresponding ground truth labels:
\begin{equation}
\mathcal{L}_{L}(\hat{\mathcal{S}}_L, \mathcal{S}_{L}) = 1- \mathcal{D}(\hat{\mathcal{S}}_L, \mathcal{S}_{L})\;,
\label{eq:largefinal}
\end{equation}
where $\mathcal{D}(\hat{\mathcal{S}}_L, \mathcal{S}_{L})$ measures the overlapping degree between two sets:
\begin{equation}
\mathcal{D}(\hat{\mathcal{S}}_L, \mathcal{S}_{L}) = \frac{\sum_{\hat{y}_{L,n}\in \hat{\mathcal{S}}_L} 2  \hat{y}_{L,n} y_{L,n}} {\sum_{\hat{y}_{L,n}\in \hat{\mathcal{S}_L}}  \hat{y}_{L,n} + \sum_{y_{L,n}\in \mathcal{S}_L} y_{L,n}}\;.
\label{eq:dice}
\end{equation}
Obviously, the loss defined by Eq. (\ref{eq:largefinal}) and (\ref{eq:dice}) focuses on overlapping error, which greatly alleviates the bias towards majority class like cross-entropy loss as well as minority class like CBCE loss. Dice loss produces serious penalty on misidentification of large hard exudate regions while slight penalty on misidentification of small hard exudate regions, which results in bias towards large hard exudate regions. 

To remedy the misidentification on small hard exudate regions, from prediction by small hard exudate biased segmentation branch, we use a re-balanced pixel sampler to sample pixels involving loss calculation. Particularly, we oversample the exudate pixels and undersample background pixels. Thus hard exudate pixels in small regions are sampled multiple times with a high confidence, which increases the penalty of misidentification on small hard exudate regions. As a result, the learning focus is shifted into small hard exudate regions. In formulation, we randomly sample $N_1$ hard exudate pixels and $N-N_1$ background pixels with replacement. We denote the sampled pixel set with $\hat{\mathcal{S}}_{S}=\{\hat{y}_{S,n}\}_{n=1}^{N}$ and the corresponding set of ground truth with $\mathcal{S}_{S}=\{y_{S,n}\}_{n=1}^{N}$ where $y_{S, n}$ is ground truth of $\hat{y}_{S, n}$. Similarly, the loss for small hard exudate biased segmentation branch is calculated as follows:
\begin{equation}
\mathcal{L}_{S}(\hat{\mathcal{S}}_S, \mathcal{S}_{S}) = 1- \mathcal{D}(\hat{\mathcal{S}}_S, \mathcal{S}_{S})\;,
\label{eq:smallfinal}
\end{equation}
where $\mathcal{D}(\hat{\mathcal{S}}_S, \mathcal{S}_{S})$ is the Dice coefficient which is computed same with Eq. (\ref{eq:dice}).

As the segmentation of hard exudates in large size is much easier than in small size, we propose an easy-to-difficult learning strategy. We adaptively modulate the losses of two branches such that the dual-branch network first learns to handle the easy task, then focuses on difficult task. At the beginning of learning, we multiply the loss of large hard exudate biased segmentation branch by a large weight $\alpha$ while the loss of small hard exudate biased segmentation branch by a small weight $1 - \alpha$ to enforce dual-branch network learn to segment hard exudate in large size. As the large hard exudate biased segmentation branch becomes more and more sophisticated in segmentation of large hard exudates, we gradually decrease the loss weight $\alpha$ and increase $1-\alpha$. In this way, the focus of dual-branch network is shifted to segmentation of small hard exudates gradually. Formally, we express our proposed dual-sampling modulated Dice loss as
\begin{equation}
\mathcal{L}_{DSM} = (1-\alpha) \cdot \mathcal{L}_{L}(\hat{\mathcal{S}}_L, \mathcal{S}_{L}) + \alpha\cdot  \mathcal{L}_{S}(\hat{\mathcal{S}}_S, \mathcal{S}_{S})\;,
\label{eq:smallfinal}
\end{equation}
where $\alpha$ is relative to learning epoch
\begin{equation}
\alpha = 1 -\left( \frac{epoch}{epoch_{max}}\right)^2\;.
\label{eq:smallfinal}
\end{equation}


\textbf{Inference.} In inference phase, the test fundus image is fed into both branches and two predictions are obtained. As both branches are equally important, we simply perform element-wise average on two predictions to obtain the final prediction.

\section{Experiments}
In this section, we first introduce the data and evaluation metrics for hard exudate segmentation, and present the implementation details, then give ablation analysis and finally make comparisons with state-of-the-arts.

\subsection{Data and Evaluation Metrics}
\textbf{Data.} In our experiments, we validate our method on two public datasets for hard exudate segmentation. The first one is the DDR dataset \cite{DDR}, which is made public for diabetic retinopathy classification, lesion segmentation and detection in 2019. To our best knowledge, DDR \cite{DDR} is the largest dataset for hard exudate segmentation. Fundus images in DDR \cite{DDR} are collected from 147 hospitals, covering 23 provinces in China and their size ranges from $1088\times1920$ to $3456\times5184$. The large variant in image size and large domain gap make the classification and segmentation on DDR \cite{DDR} more challenge. Specific to lesion segmentation, DDR \cite{DDR} provides 757 fundus images with pixel-level annotation, among which 383 images are for training, 149 for validation and 225 for testing. The other is IDRiD \cite{idrid2018, idrid2020}, provided by a grand challenge on “Diabetic Retinopathy – Segmentation and Grading” in 2018. It provides 81 fundus images size of $4288\times2848$ with pixel-level hard exudate annotations, among which 54 images are used for training and 27 for testing. All of those images are acquired from an eye clinic in India. 

\textbf{Evaluation Metrics.} We evaluate segmentation methods at both pixel-level and region-level. With regard to pixel-level metrics, following \cite{DDR} and \cite{idrid2020}, Intersection of Union (IoU) and Area Under Precision-Recall Curve (AUPR) are adopted. We also adopt F-score for evaluation, which is harmonic mean of sensitivity ($SN$) and positive predicted value ($PPV$). 

With regard to region-level metrics, we follow \cite{zhang2014exudate, liu2017location, binloss} and re-define true positive (TP), false positive (FP) and false negative (FN). We denote predicted hard exudate connected component set with $\hat{C}=\{\hat{C}_1,\cdots, \hat{C}_N \}$ and ground truth hard exudate connected component set with $C = \{C_1, \cdots, C_M\}$. A pixel is defined as a TP if, and only if, it belongs to:
\begin{equation}
\{\hat{C} \cap C\} \cup \left\{ \hat{C}_i \bigg| \frac{|\hat{C}_i \cap C|}{|\hat{C}_i|} > \sigma \right\} \cup \left\{ C_i \bigg| \frac{|\hat{C}_i \cap C|}{|C_i|} > \sigma \right\}\;,
\end{equation}
where $|\cdot|$ accounts the cardinality and $\sigma\in [0,1]$ is the overlap ratio threshold. The larger $\sigma$ is, more rigorous the condition that a pixel is treated as a TP is. A pixel is considered as an FP if, and only if it belongs to
\begin{equation}
\{\hat{C}_i | \hat{C}_i \cap C = \emptyset \}
\cup \left\{ \hat{C}_i \cap \bar{C} \bigg| \frac{|\hat{C}_i \cap C|}{|\hat{C}_i|} \leq \sigma \right\}\;,
\end{equation}
where $\bar{C}$ is complementary set to $C$. A pixel is considered as an FN if, and only if it belongs to 
\begin{equation}
\{C_i | \hat{C} \cap C_i = \emptyset \}
\cup \left\{ C_i \cap \bar{\hat{C}} \bigg| \frac{|C_i \cap \bar{\hat{C}}|}{|C_i|} \leq \sigma \right\}\;,
\end{equation}
where $\bar{\hat{C}}$ is complementary set to $\hat{C}$. Rest pixels are considered as TNs. The region-level F-score is defined as 
\begin{equation}
F_{\sigma} = \frac{2\times SN_\sigma \times PPV_\sigma}{SN_\sigma + PPV_\sigma}\;,
\end{equation}
where $SN_\sigma$ is sensitivity defined as $SN_\sigma=\frac{TP}{TP+FN}$ and $PPV_\sigma$ is positive predictive value defined as $PPV_\sigma=\frac{TP}{TP+FP}$. In our experiment, $F_{0.2}$, $F_{0.35}$, $F_{0.5}$, $F_{0.65}$ and $F_{0.8}$ are reported.

\subsection{Implementation Details}
\textbf{Data Preprocessing and Augmentation.} For images in DDR \cite{DDR}, we first crop the bounding box of field of view, then for cropped boxes whose short side is less than 1024, we enlarge the short side to equal length of long side via zero padding. Finally, we resize them to $1024\times1024$. For images in IDRiD \cite{idrid2018, idrid2020}, we directly resize images to $1440\times960$. Following \cite{guo2019l-seg, binloss}, on both datasets, two tricks are adopted to augment the training data: rotation (90$^{\circ}$, 180$^{\circ}$ and 270$^{\circ}$) and flipping (horizontal and vertical).

\textbf{Experimental Setting.} We build three variants of dual-branch segmentation network based on PSPNet \cite{PSPNet}, Deeplabv3 \cite{Deeplabv3} and HED \cite{HED}. We call them dual-PSPNet, dual-Deeplabv3 and dual-HED respectively and implement them within PyTorch framework. We initialise the parameters in backbones with the pre-trained model on ImageNet \cite{ImageNet} and the parameters associated with classifiers with Gaussian distribution with zero mean and standard deviation of 0.01. SGD is used for parameter optimisation. Hyper-parameters include: initial learning rate(0.03 poly policy with power of 0.9), weight decay (0.0005), momentum (0.9), batch size (2) and iteration epoch (100 on DDR \cite{DDR} and 40 on IDRiD \cite{idrid2018}). Sample rate $N_1/N$ in re-balanced pixel sampler is set to 0.5. The models are trained on GeForce RTX 2080 Ti with two GPUs. 

\subsection{Ablation Study}
\label{subsction:AblationStudy}
\textbf{Effect of Sample Rate.} We take dual-PSPNet as example and first explore the effectiveness of sample rate $N_1/N$ of re-balanced pixel sampler. We set it to 0.25, 0.5, 0.75 and reversed class frequency and train dual-PSPNet on DDR \cite{DDR} training set separately. Results on test set are reported in Table. \ref{AblationSampleRate}.  They show that our dual-PSPNet network achieves best when sample rate $N_1/N$ is set to 0.5. In what follows, except for extra illustration,  $N_1/N = 0.5$  is the default setting.

\begin{table*}[tbh!]
	\caption{Effect of sample rate of re-balanced sampler in our Dual PSPNet on DDR \cite{DDR} test set. $SN_{pixel}$, $PPV_{pixel}$ and $F_{pixel}$ are pixel-level sensitivity, positive predicted value and F-score.}
	\centering
	\begin{tabular}{|c|c|c|c|c|c|}
		\hline
		$N_1/N$ & $SN_{pixel}$ & $PPV_{pixel}$  & $F_{pixel}$ & $IoU$ & $AUPR$\\	\hline
		0.25 &	0.5863 & 	\textbf{0.5706} &	0.5784 	& 0.4069 & 0.5468 \\
		0.5	& \textbf{0.6077} &	0.5582 &	\textbf{0.5819}  & \textbf{0.4103} & \textbf{0.5587} \\
		0.75 & 0.5947  & 0.5628 & 0.5783  & 0.4068 & 0.5491 \\
		reverse class frequency &0.5673 & 0.5547 & 0.5609 & 0.3898 & 0.5202 \\\hline
	\end{tabular}	
	\label{AblationSampleRate}
\end{table*}

\begin{table*}[tbh!]
	\caption{Comparisons of single branch PSPNet \cite{PSPNet}, Deeplabv3 \cite{Deeplabv3} and HED \cite{HED} with CBCE loss and Dice loss and our dual-branch ones with DSM loss on DDR \cite{DDR} test set.}
	\centering
	\begin{tabular}{|c||c|c|c||c||c||c|c|c|c|c|}
		\hline
		Method & $SN_{pixel}$ &  $PPV_{pixel}$ & $F_{pixel}$ & $IoU$ & $AUPR$ & $F_{\sigma=0.2}$ & $F_{\sigma=0.35}$ & $F_{\sigma=0.5}$ & $F_{\sigma=0.65}$ & $F_{\sigma=0.8}$\\		
		\hline
		PSPNet + CBCE & \textbf{0.7014} & 0.2934 & 0.4137 & 0.2608  & 	0.4906 & 0.7441  & 0.5880 & 0.4917 & 0.4587 & 0.4257\\
		PSPNet + Dice & 0.4425 & \textbf{0.7372} & 0.5530 & 0.3822  & 	0.4730 & 0.8640 & 0.7790 & 0.6954 & \textbf{0.6467} & 0.5902 \\
		Dual PSPNet + DSM & 0.6077 & 0.5582 & \textbf{0.5819} & \textbf{0.4103}  & \textbf{0.5587} & \textbf{0.8887} & \textbf{0.8454} & \textbf{0.7543} & 0.6428 & \textbf{0.6034} \\ \hline
		
		Deeplabv3 + CBCE & \textbf{0.6557} & 0.3431 & 0.4505 & 0.2907 &  	0.4881 & 0.7968 & 0.6652 & 0.5564 & 0.4856 & 0.4617 \\
		Deeplabv3 + Dice & 0.4210 & \textbf{0.7395} & 0.5366 & 0.3667 & 	0.4495 & 0.8310 & 0.7341 & 0.6768 & 0.6334 & 0.5754 \\
		{Dual Deeplabv3+DSM} & 0.5170 & 0.6569 & \textbf{0.5786} & 	\textbf{0.4071} & \textbf{0.5701} & \textbf{0.8811} & \textbf{0.8372} & \textbf{0.7542} & \textbf{0.6562} & \textbf{0.6053} \\\hline
		
		HED + CBCE & \textbf{0.7302} & 0.2617 & 0.3853 & 0.2386 & \textbf{0.5342} & 0.6823 & 0.5672 & 0.4705 & 0.4317 & 0.3959 \\
		HED + Dice  & 0.4899 & \textbf{0.7032} & 0.5775 & 0.4060 &  	0.4775 & 0.8451 & 0.7807 & 0.7107 & \textbf{0.6776} & \textbf{0.6164} \\
		Dual HED + DSM & 0.6006 & 0.5714 & \textbf{0.5856} & \textbf{0.4140}  & 	0.5294 & \textbf{0.8809} & \textbf{0.8402} & \textbf{0.7205} & 0.6419 & 0.6099 \\\hline
	\end{tabular}	
	\label{AblationDDR}
\end{table*}

\begin{table*}[tbh!]
	\caption{Comparisons of single branch PSPNet \cite{PSPNet}, Deeplabv3 \cite{Deeplabv3} and HED \cite{HED} with CBCE losse and Dice loss and  our dual-branch ones with DSM loss on IDRiD \cite{idrid2018} test set. }
	\centering
	\begin{tabular}{|c||c|c|c||c||c||c|c|c|c|c|}
		
		\hline
		Method & $SN_{pixel}$ &  $PPV_{pixel}$ & $F_{pixel}$ & $IoU$ & $AUPR$ & $F_{\sigma=0.2}$ & $F_{\sigma=0.35}$ & $F_{\sigma=0.5}$ & $F_{\sigma=0.65}$ & $F_{\sigma=0.8}$\\		
		\hline
		PSPNet + CBCE & \textbf{0.9632} & 0.3361 &	0.4983 & 0.3318 & 0.7475 & 0.8254 &	0.7036 	&0.6268 	&0.5463 	&0.5024  \\
		PSPNet + Dice	& 0.7358 &	\textbf{0.7819} & 0.7582 & 0.6106& 0.7878 &0.9388 	&0.9297 	&0.9160 &	0.8924 &	0.8304 \\
		Dual PSPNet + DSM & 0.7748  & 0.7572 &
		\textbf{0.7659} & \textbf{0.6206} & \textbf{0.7977} &	\textbf{0.9469} &	\textbf{0.9386} &	\textbf{0.9241} &	\textbf{0.8946} &	\textbf{0.8351} \\\hline	
		
		Deeplabv3 + CBCE & 	\textbf{0.9492} 	& 0.3518  &	0.5134 & 0.3453 & 0.7335 & 	0.8532 & 0.7410 & 0.6467 & 	0.5244 & 0.5190\\  
		Deeplabv3 + Dice & 0.7527 	& \textbf{0.7690} &	0.7607 & 0.6139 & \textbf{0.7891} & 	0.9462 & 0.9344 & 0.9198 & 0.8925 & 0.8320 \\
		{Dual Deeplabv3+DSM} & 0.7686 &	0.7621 & \textbf{0.7653}  & \textbf{0.6198} & 0.7890 & 	\textbf{0.9478}& 	\textbf{0.9374}& 	\textbf{0.9240}& 	\textbf{0.8973}& 	\textbf{0.8347}  \\\hline
		
		HED + CBCE & \textbf{0.9019} & 	0.4065  & 0.5604 & 0.3893 & 0.7740 & 0.9013 & 	0.8219 & 0.7419 &  0.5967 & 	0.5664 \\
		HED + Dice & 0.7211 &	0.7640   & 0.7419 & 0.5897 & 0.7894 & 	0.9429 & 	0.9325 & 	\textbf{0.9202} & 	0.8698 & 	0.8038\\
		Dual HED + DSM & 0.7630 & \textbf{0.7739}  & \textbf{0.7684} & \textbf{0.6239} & \textbf{0.8296} & 	\textbf{0.9446} & 	\textbf{0.9348} & 	0.9184 & 	\textbf{0.8971} & 	\textbf{0.8353} \\\hline	
	\end{tabular}	
	\label{AblationIDRiD}
\end{table*}

\begin{figure*}[tbh!]
	\centering
	\includegraphics[width=\textwidth]{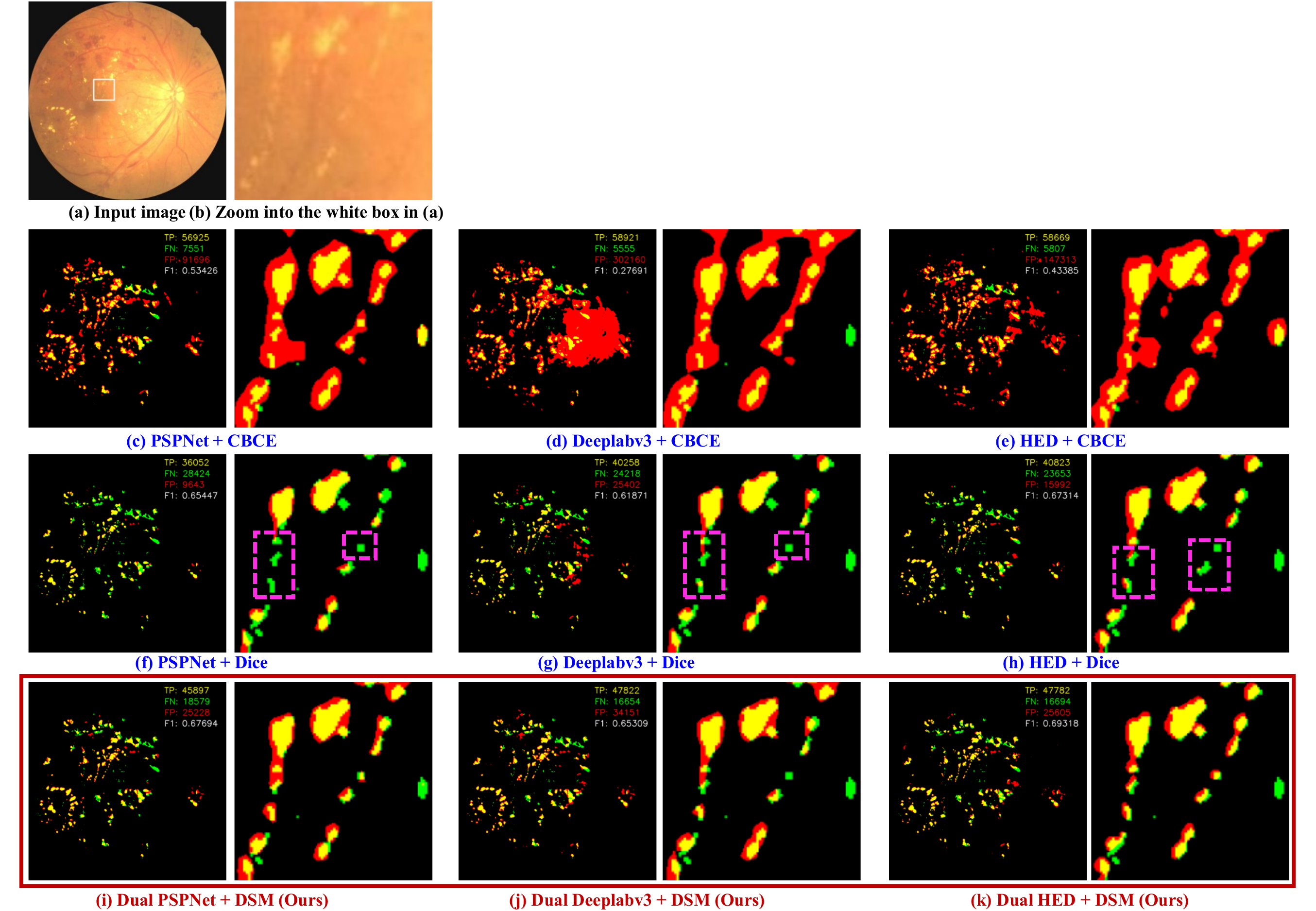}
	\caption{Visual comparisons of proposed dual-branch networks with DSM loss to single branch networks with CBCE loss and Dice loss on DDR \cite{DDR}. The single branch networks are PSPNet \cite{PSPNet}, Deeplabv3 \cite{Deeplabv3} and HED \cite{HED}. Pixels in yellow are exudate pixels that are correctly classified. Pixels in red are background pixels that are wrongly classified as exudate pixels. Pixels in green are exudate pixels that are wrongly classified as background pixels. Dashed magenta boxes highlight hard exudates that are misidentified.}
	\label{ablation_study_ddr1}
\end{figure*}

\begin{figure*}[tbh!]
	\centering
	\includegraphics[width=\textwidth]{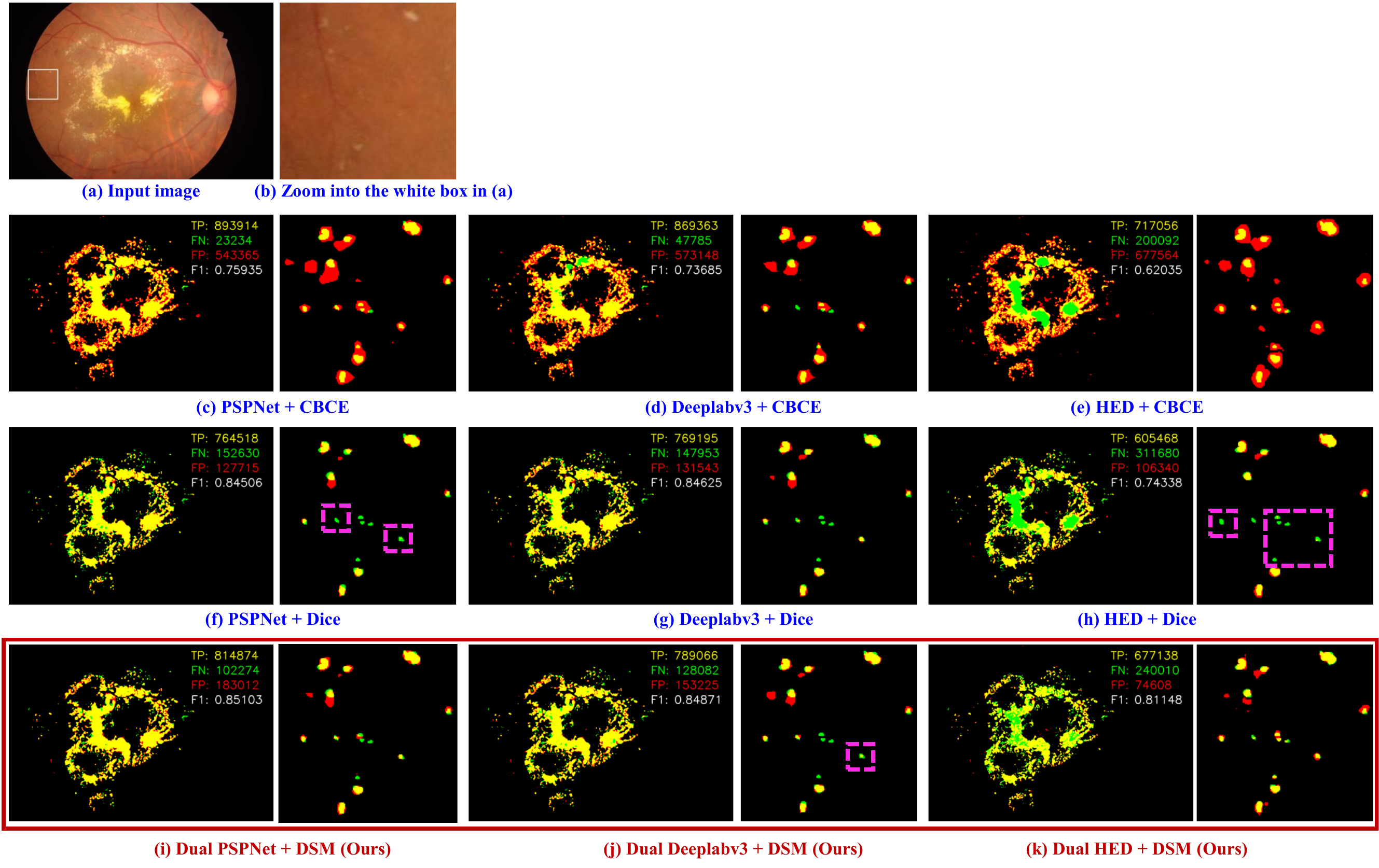}
	\caption{Visual comparisons of proposed dual-branch networks with DSM loss to single branch networks with CBCE loss and Dice loss on IDRiD \cite{idrid2018}. The single branch networks are PSPNet \cite{PSPNet}, Deeplabv3 \cite{Deeplabv3} and HED \cite{HED}. Pixels in yellow are exudate pixels that are correctly classified. Pixels in red are background pixels that are wrongly classified as exudate pixels. Pixels in green are exudate pixels that are wrongly classified as background pixels. Dashed magenta boxes highlight hard exudates that are misidentified.}
	\label{ablation_study_idirid1}
\end{figure*}

\textbf{Ablation Study on Different Losses.} To evaluate dual-branch network, we conduct experiments on three state-of-the-art dense classification methods PSPNet \cite{PSPNet}, Deeplabv3 \cite{Deeplabv3} and HED \cite{HED} with several settings, including single branch network with CBCE loss, single branch network with Dice loss \cite{diceloss} and our proposed dual-branch network with DSM loss on both DDR \cite{DDR} and IDRiD \cite{idrid2018}. 

Table. \ref{AblationDDR} reports the results on DDR \cite{DDR}. Obviously, single branch networks with Dice loss \cite{diceloss} are much more sophisticated in hard exudate segmentation than with CBCE loss. This is because that CBCE loss re-weights costs according to reverse class frequency. It over-weights costs on hard exudate pixels and under-weights costs on background pixels, which makes the network be biased in favour of hard exudates. Thus the pixel-level sensitivity is very high while the PPV is very low. The harmonic mean i.e. pixel-level F-score is low. On the contrary, single network branches trained with Dice loss improve the pixel-level PPV significantly but inferior sensitivity. Nevertheless, final pixel-level F-score has been greatly improved. Compared with single branch networks with Dice loss, our dual branch networks with proposed DSM loss further improve the performance. In terms of IoU, our dual-branch ones with DSM loss are superior to the single branch ones. In terms of AUPR, our dual-branch ones outperform single branch ones except for the dual HED. In terms of region-level evaluation metrics, from Table. \ref{AblationDDR}, we can see that dual Deeplabv3 \cite{Deeplabv3} with DSM loss consistently achieves better than single branch ones. Our dual PSPNet with DSM outperforms the single branch PSPNet \cite{PSPNet} with CBCE loss and Dice loss  except for $\sigma=0.65$. Our dual HED  with DSM outperforms the single branch HED \cite{HED} with Dice loss when $\sigma$ is small. As $\sigma$ is larger than 0.5, our dual HED with DSM loss achieve inferior regional F-score to single branch network trained with Dice loss. The possible reason is that our dual HED with DSM segment small sized hard exudates coarsely and background pixels around small sized hard exudates are misidentified. When $\sigma$ is small, those misidentified background pixels are treated as true positives. Thus our dual HED with DSM loss achieve high regional F-scores than single branch network with Dice loss. When $\sigma$ increases, those misidentified background pixels are considered as false positives, thus the regional F-scores of ours are lower than single branch one with Dice loss. On IDRiD \cite{idrid2018}, from Table. \ref{AblationIDRiD} we can see that our dual-branch networks with DSM outperform single branch networks with CBCE and Dice loss except for Dual HED with DSM when $\sigma$ is 0.5.

In terms of number of parameters, parameters in two branches with PSPNet \cite{PSPNet} and Deeplabv3 \cite{Deeplabv3}  are almost 1.5 times of those in single branches while two branches with HED \cite{HED} is 1.9 times of that in single branch.

Visual comparisons on DDR \cite{DDR} and IDRiD \cite{idrid2018} of our dual-branch networks with DSM loss to single branch networks with CBCE and Dice losses are provided in Fig. \ref{ablation_study_ddr1} and Fig. \ref{ablation_study_idirid1} respectively. We can see that single networks with CBCE loss are prone to misidentify background pixels around hard exudates. Single networks with Dice loss are prone to misidentify small size hard exudates. Our dual-branch networks work better than single branch networks.

\subsection{Comparison with State-of-the-arts}
On DDR \cite{DDR}, we compare our three dual-branch networks with three deep learning based methods: Deeplabv3+ \cite{Deeplabv3plus}, DNL \cite{yin2020disentangled} and SPNet \cite{hou2020strip}. For DNL \cite{yin2020disentangled} and SPNet \cite{hou2020strip}, we provide results with two losses: CBCE and Dice loss. All these methods are originally designed for natural scene image segmentation. Table. \ref{rstDDR} reports the results. We note that results of Deeplabv3+ \cite{Deeplabv3plus} in the first row are provided by \cite{DDR} and the rest are obtained by fine-tuning on DDR \cite{DDR}. As shown in Table. \ref{rstDDR}, our dual-branch networks achieve superior performance in terms of both pixel-level and region-level metrics.

On IDRiD \cite{idrid2018}, we compare our dual-branch network with five deep learning based  methods: DNL \cite{yin2020disentangled}, SPNet \cite{hou2020strip}, L-seg \cite{guo2019l-seg}, LWENet \cite{LWENet} and Bin loss \cite{binloss}. For LWENet \cite{LWENet} and L-seg \cite{guo2019l-seg}, the predicted binary masks are provided by authors. With them, $IoU$, $F_{pixel}$ and the region-level metrics are computed. For the rest methods, results are obtained by fine-tuning. Table. \ref{rstIDRiD} reports the results. We can see that our dual-branch network achieves superior results than compared methods.

\begin{table*}[tbh!]
	\caption{Comparison of other segmentation methods on DDR \cite{DDR} test set. Results of Deeplabv3+ \cite{Deeplabv3plus} are directly borrowed from \cite{DDR}.}
	\centering
	\begin{tabular}{|c||c|c|c||c|c|c|c|c|}
		\hline
		Method  & $IoU$ & $F_{pixel}$ & $AUPR$ & $F_{\sigma=0.2}$ & $F_{\sigma=0.35}$ & $F_{\sigma=0.5}$ & $F_{\sigma=0.65}$ & $F_{\sigma=0.8}$\\		
		\hline
		Deeplabv3+ \cite{Deeplabv3plus} & 0.3118 &- &-	&- 	&- 	&- 	&- 	&-\\
		DNL \cite{yin2020disentangled} + CBCE & 0.1643  & 	0.2822  & 	0.5125 & 0.4923  & 0.3617  & 	0.3308  & 	0.3186  & 	0.2957 \\	
		DNL \cite{yin2020disentangled} + Dice & 0.3862 & 0.5572 & 0.4854 &			0.8683 & 0.8116 & 0.7026 &	0.5970 & 0.5611 \\
		SPNet \cite{hou2020strip} + CBCE & 0.1034 & 0.1874 & 0.5034 &			0.2856 & 0.2395 & 0.2264 & 0.2073 & 0.1942  \\
		SPNet \cite{hou2020strip} + Dice & 0.3089 & 0.4720 & 0.3748 &			0.7469 & 0.7151 & 0.6291 &	0.5395 & 0.4948 \\	\hline
		{Dual PSPNet + DSM  (Ours)} &	0.3822  & 0.5530 &  	0.4730 & 0.8640 & 0.7790 & 0.6954 & {0.6467} & 0.5902
		\\
		{Dual Deeplabv3 + DSM  (Ours)}& {0.4071} &	{0.5786} & {0.5701} & {0.8811} & {0.8372} & {0.7542} & {0.6562} & {0.6053} \\
		Dual HED + DSM (Ours) &  {0.4140}  & {0.5856} &	{0.5294} & {0.8809} & {0.8402} & {0.7205} & 0.6419 & 0.6099
		\\\hline				
	\end{tabular}
	
	\label{rstDDR}
\end{table*}

\begin{table*}[tbh!]
	
	\caption{Comparison of other segmentation methods on IDRiD \cite{idrid2018} test set.}
	\centering
	\begin{tabular}{|c||c|c||c||c|c|c|c|c|}
		\hline
		\textbf{Method} & ${IoU}$ & ${F_{pixel}}$ & $AUPR$ & ${F_{\sigma=0.2}}$ & ${F_{\sigma=0.35}}$ & ${F_{\sigma=0.5}}$ & ${F_{\sigma=0.65}}$ & ${F_{\sigma=0.8}}$\\		
		\hline
		DNL\cite{yin2020disentangled} + CBCE & 0.2304 &	0.3745 & 0.6807 &			0.6709 &	0.5753 &	0.4671 &	0.3771 &	0.3760 \\	
		DNL\cite{yin2020disentangled} + Dice & 0.5715 & 0.7273 & 0.6891 &		0.9413 & 0.9283 &	0.9062 & 0.8698 & 0.8041 \\	
		SPNet \cite{hou2020strip} + CBCE & 0.2143 & 0.3530 & 0.6223 &			0.6412 &	0.5384 &	0.4420 &	0.3568 &	0.3552  \\
		SPNet \cite{hou2020strip} + Dice & 0.4923 &	0.6598 &	0.6064 	&		0.9258 &	0.8883 &	0.8642 &	0.8145 &	0.7147  \\
		L-seg \cite{guo2019l-seg} & 0.5909 & 0.7429 & - &0.9508 &	0.9417 &	0.9271 &	0.8754 &	0.8038 
		\\ 
		LWENet \cite{LWENet} & 0.5226 & 0.6865 & - &0.9191 & 0.8942& 0.8582& 0.8179& 0.7160\\
		Bin loss \cite{binloss} &
		0.3582 &	0.5275&	0.7047 &0.9070& 	0.8056 &	0.6890 &	0.5400 &	0.5319 		\\\hline
		Dual PSPNet + DSM (Ours) & {0.6206} &	{0.7659} & 0.7977&	{0.9469} &	{0.9386} &	{0.9241} &	{0.8946} &	{0.8351} \\
		
		Dual Deeplabv3 + DSM  (Ours) &  {0.6198} &  {0.7653}  & 0.7890 & 	{0.9478}& 	{0.9374} & 	{0.9240}& 	{0.8973}& 	{0.8347}  \\
		
		Dual HED + DSM & {0.6239} & {0.7684} &  {0.8296} & 	{0.9446} & 	{0.9348} & 	0.9184 & 	{0.8971} & 	{0.8353}\\	\hline		
	\end{tabular}
	
	\label{rstIDRiD}
\end{table*}

\begin{figure*}[t]
	\centering
	\includegraphics[width=\textwidth]{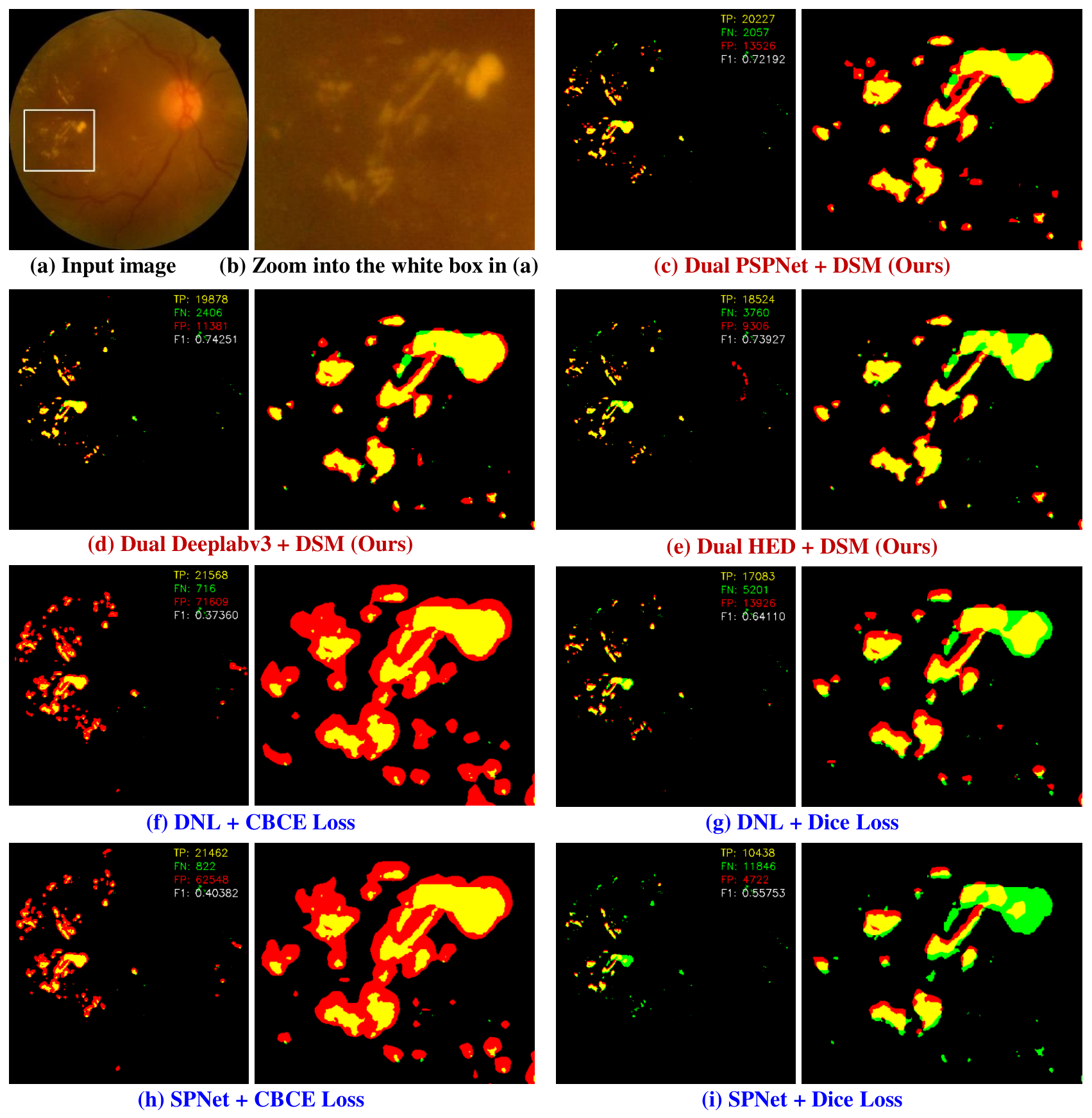}
	\caption{Visual comparisons with DNL \cite{yin2020disentangled}, SPNet \cite{hou2020strip} with CBCE loss and Dice loss on DDR \cite{DDR}. Pixels in yellow are exudate pixels that are correctly classified. Pixels in red are background pixels that are wrongly classified as exudate pixels. Pixels in green are exudate pixels that are wrongly classified as background pixels.}
	\label{VisualComparisonDDR}
\end{figure*}

\begin{figure*}[t]
	\centering
	\includegraphics[width=0.96\textwidth]{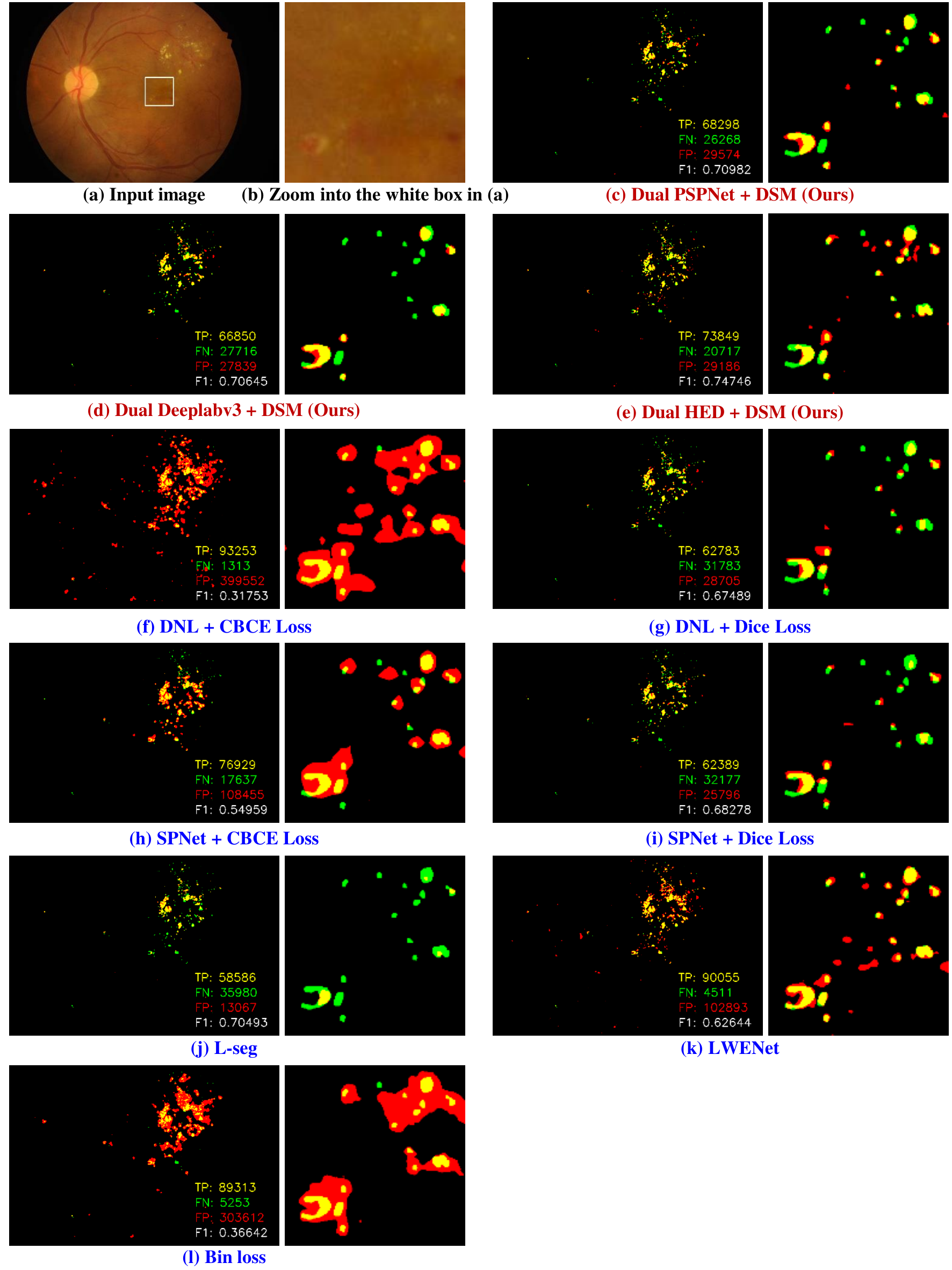}
	\caption{Visual comparisons of with DNL \cite{yin2020disentangled}, SPNet \cite{hou2020strip} with CBCE loss and Dice loss, L-seg \cite{guo2019l-seg}, LWENet \cite{LWENet} and Bin loss \cite{binloss} on IDRiD \cite{idrid2018}. Pixels in yellow are exudate pixels that are correctly classified. Pixels in red are background pixels that are wrongly classified as exudate pixels. Pixels in green are exudate pixels that are wrongly classified as background pixels.}
	\label{VisualComparisonIDRiD}
\end{figure*}

Visual comparisons between previous methods and ours on DDR \cite{DDR} and IDRiD \cite{idrid2018} are performed. In Fig. \ref{VisualComparisonDDR}, the segmentation results by our dual-branch networks based on PSPNet \cite{PSPNet}, Deeplabv3 \cite{Deeplabv3} and HED \cite{HED}, DNL \cite{yin2020disentangled} with CBCE loss and Dice loss, SPNet \cite{hou2020strip} with CBCE loss and Dice loss are shown. We can see that (1) the compared DNL \cite{yin2020disentangled} and SPNet \cite{hou2020strip} with CBCE loss are hard exudate biased and prone to misidentify background pixels as hard exudate pixels; (2) SPNet \cite{hou2020strip} with Dice loss are background biased and prone to misidentify hard exudate pixels as background pixels; (3) Our dual-branch networks achieve better than them. In Fig. \ref{VisualComparisonIDRiD}, the segmentation results by ours, DNL \cite{yin2020disentangled} SPNet \cite{hou2020strip},  L-seg \cite{guo2019l-seg}, LWENet \cite{LWENet}, and Bin loss \cite{binloss} are shown. Similarly, we can see that (1) DNL \cite{yin2020disentangled} and SPNet \cite{hou2020strip} with CBCE loss are hard exudate biased while with Dice loss are background biased; (2) L-seg \cite{guo2019l-seg} is background biased while LWE \cite{LWENet} and Bin loss \cite{binloss} are hard exudate biased; (3) ours achieve better than the compared methods.

\newpage
\section{Conclusion}
In this paper, we propose dual-branch network to address the issues of extreme class imbalance and enormous variation in size across target regions for segmentation of hard exudate in colour fundus images. Our dual-branch network uses two branches with partial weight sharing to learn representations and classifiers for hard exudates in different sizes. It is trained with the proposed dual-sampling modulated Dice loss, which enables dual-branch network to first learn to segment large hard exudates then small hard exudates. Experimental results on two public datasets for hard exudate segmentation demonstrate that our dual-branch network outperforms existing segmentation networks with both CBCE loss and Dice loss.

\bibliographystyle{aaai21}

\begin{small}
	\bibliography{ref}
\end{small}

\end{document}